\documentclass[runningheads]{llncs}
\usepackage[T1]{fontenc}
\usepackage{graphicx,verbatim}
\usepackage{booktabs}
\usepackage{multirow}
\usepackage{array}
\usepackage[table]{xcolor}
\usepackage{amsmath}
\usepackage{amssymb}
\usepackage{bbding}
\usepackage{xcolor}
\definecolor{lightgray}{gray}{0.92}
\usepackage{color}
\usepackage{hyperref}

\begin{document}
\title{Directed Ordinal Diffusion Regularization for Progression-Aware Diabetic Retinopathy Grading}
%\titlerunning{Abbreviated paper title}
% If the paper title is too long for the running head, you can set
% an abbreviated paper title here
%
% \begin{comment}  %% Removed for anonymized MICCAI submission
\author{Huangwei Chen\inst{1,2,3}$^{\dagger}$ \and
Junhao Jia\inst{1,3}$^{\dagger}$ \and
Ruocheng Li\inst{1} \and
Cunyuan Yang\inst{1} \and
Wu Li\inst{1} \and
Xiaotao Pang\inst{2} \and
Yifei Chen\inst{4} \and
Haishuai Wang\inst{1}\textsuperscript{\smash{(\Envelope)}} \and
Jiajun Bu\inst{1} \and
Lei Wu\inst{1,2}\textsuperscript{\smash{(\Envelope)}}
}
\authorrunning{F. Author et al.}
% First names are abbreviated in the running head.
% If there are more than two authors, 'et al.' is used.
%
\institute{
Zhejiang Key Laboratory of Accessible Perception and Intelligent Systems, College
of Computer Science and Technology, Zhejiang University
\and Hangzhou Pujian Medical Technology Co., Ltd
\and Hangzhou Dianzi University \quad \textsuperscript{4} Tsinghua University\\
\email{\{haishuai.wang, shenhai1895\}@zju.edu.cn}}

% \end{comment}

% \author{Anonymized Authors}  %% Added for anonymized MICCAI submission
% \authorrunning{Anonymized Author et al.}
% \institute{Anonymized Affiliations \\
%     \email{email@anonymized.com}}
  
\maketitle              % typeset the header of the contribution
\begingroup
\renewcommand{\thefootnote}{}
\footnotetext{$^{\dagger}$~Equal contribution; \Envelope~corresponding author.}
\endgroup
\begin{abstract}
Diabetic Retinopathy (DR) progresses as a continuous and irreversible deterioration of the retina, following a well-defined clinical trajectory from mild to severe stages. However, most existing ordinal regression approaches model DR severity as a set of static, symmetric ranks, capturing relative order while ignoring the inherent unidirectional nature of disease progression. As a result, the learned feature representations may violate biological plausibility, allowing implausible proximity between non-consecutive stages or even reverse transitions. To bridge this gap, we propose Directed Ordinal Diffusion Regularization (D-ODR), which explicitly models the feature space as a directed flow by constructing a progression-constrained directed graph that strictly enforces forward disease evolution. By performing multi-scale diffusion on this directed structure, D-ODR imposes penalties on score inversions along valid progression paths, thereby effectively preventing the model from learning biologically inconsistent reverse transitions. This mechanism aligns the feature representation with the natural trajectory of DR worsening. Extensive experiments demonstrate that D-ODR yields superior grading performance compared to state-of-the-art ordinal regression and DR-specific grading methods, offering a more clinically reliable assessment of disease severity. Our code is available on \href{https://github.com/HovChen/D-ODR}{https://github.com/HovChen/D-ODR}.

\end{abstract}
\section{Introduction}
Diabetic retinopathy (DR) is a leading cause of preventable blindness among working-age adults, and large-scale screening commonly relies on grading retinal fundus photographs into discrete severity categories~\cite{cheung2010DiabeticRetinopathy,ruamviboonsuk2022realtime,arcadu2019deeplearning}. This grading task is frequently formulated as ordinal prediction due to the inherent ordering of the labels~\cite{toledo2022dqor}. In practice, severity grades represent disease states along a temporal trajectory. Mild instances of DR may remain stable or progress, and the transition dynamics are typically asymmetric~\cite{maturi2022four}. Consequently, the estimation of severity should extend beyond preserving ordinal rank and incorporate the direction of progression inherent to the pathology~\cite{dai2024deepdr}.

Most existing learning formulations overlook this directional constraint~\cite{davor2023constrained}. Methods based on regression, ordinal classification, and ranking typically enforce symmetric consistency by penalizing disagreement between two samples without verifying whether the relationship aligns with the disease course~\cite{pitawela2025cloc,wang2023ord2seq}. Under such symmetric supervision, constraints propagate equally from severe to mild cases and vice versa~\cite{li2021poes}. Consequently, the training objective may favor numerically smooth yet clinically inconsistent solutions.

We assert that ordinal supervision for DR should be intrinsically directional~\cite{shaik2025ordinal}. A valuable inductive bias is to encode forward-biased progression dynamics, where pairwise relationships utilized for supervision reflect feasible worsening trajectories under typical screening settings. From this perspective, the representation space should admit a manifold where transitions are permitted only forward in severity, and the learning signal should enforce monotonicity along these forward transitions. Importantly, reverse relations should not impose constraints, as they correspond to biologically implausible trajectories and can introduce conflicting gradients that obscure the semantics of severity.

Motivated by this perspective, we propose Directed Ordinal Diffusion Regularization (D-ODR), integrating progression awareness into ordinal learning frameworks. During training, D-ODR constructs a directed graph within each mini-batch, establishing connections between feature-neighboring instances exclusively when the destination instance represents an equal or more advanced disease stage. The optimization objective enforces the predicted continuous severity scores to monotonically increase along these valid forward paths, which ensures that the learning signal strictly follows clinically valid trajectories.

Conceptually, the proposed approach models predicted severity as a scalar field over a directed manifold mirroring disease evolution. By propagating ordinal structure through multi-step diffusion, the regularizer captures extended progression chains while remaining anchored in local feature similarity~\cite{jia2026geoproto}. This module operates only during training, introducing zero computational overhead at inference. Evaluations across multiple public datasets and backbones show D-ODR yields consistent improvements. The results validate that respecting disease progression irreversibility serves as a practical principle for DR grading.

\section{Methods}

\subsection{Model Overview}
\begin{figure}
    \centering
    \includegraphics[width=0.95\linewidth]{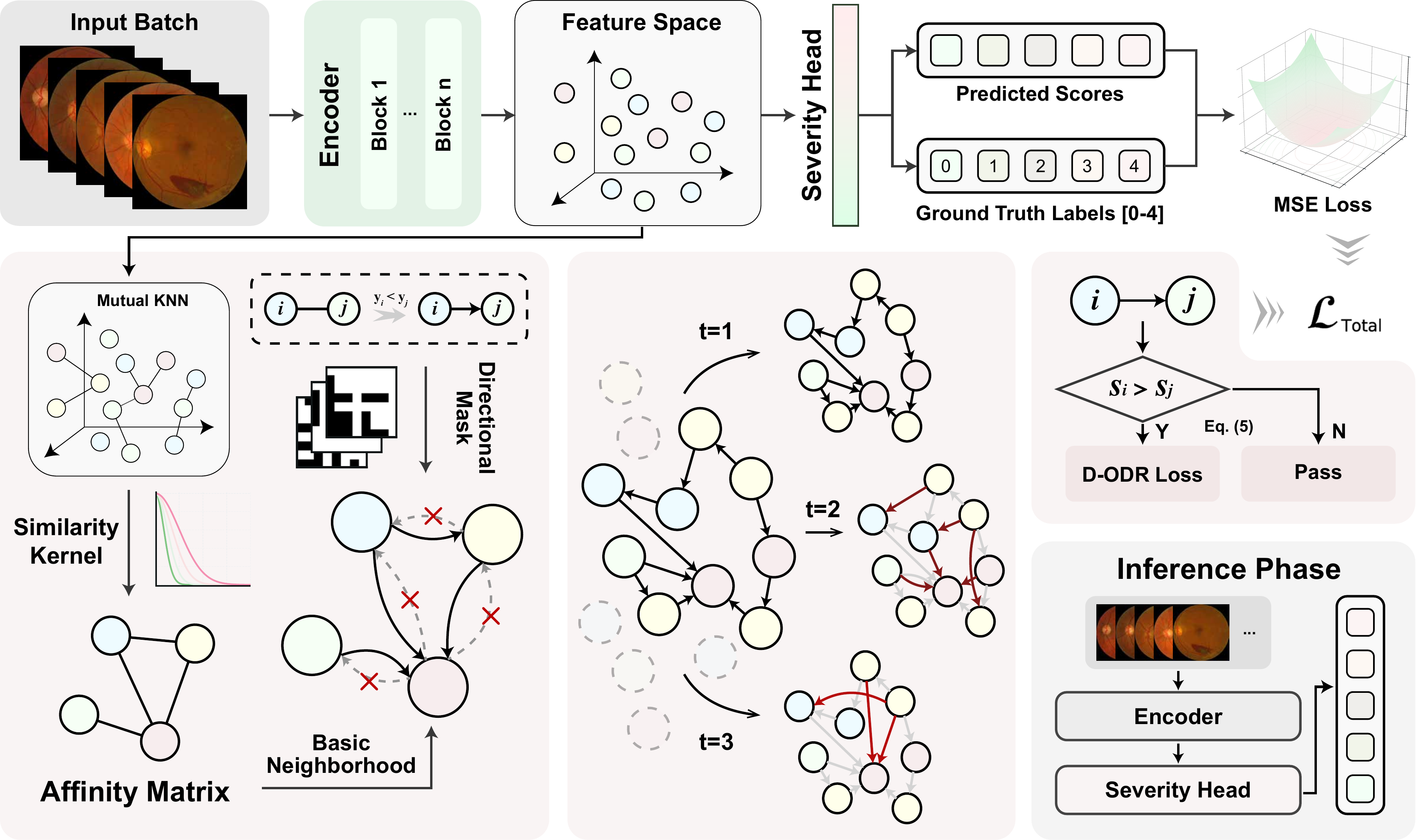}
    \caption{Overview of the proposed D-ODR framework. The model constructs a directed graph in the feature space using a directional mask and applies multi-scale diffusion to capture extended progression chains, therefore penalizing score inversion along valid forward trajectories. The regularization module is discarded during inference, ensuring no extra computational cost.}
    \label{fig:pipeline}
\end{figure}

Given a retinal fundus dataset $\mathcal{D}=\{(x_i,y_i)\}_{i=1}^{N}$ with ordinal severity labels $y_i\in\{0,1,2,3,4\}$, our objective is to learn a model $f_\theta$ that predicts a continuous severity score. As illustrated in Fig.~\ref{fig:pipeline}, $f_\theta$ comprises a feature encoder $g_\theta$ and a severity prediction head $h_\phi$. For each input image $x_i$, the encoder extracts a $d$-dimensional representation $z_i=g_\theta(x_i)\in\mathbb{R}^d$, which the head subsequently maps to a continuous score $s_i=h_\phi(z_i)\in\mathbb{R}$.

This architecture is governed by a fundamental clinical prior: DR progression is effectively irreversible, typically evolving from mild to severe stages. Consequently, disease samples in the representation space are expected to lie on a manifold with a directional constraint, where local transitions are valid only along paths of non-decreasing severity. To explicitly align model predictions with this forward progression, we introduce D-ODR. Within each training mini-batch, D-ODR constructs a label-driven directed graph in the feature space, enforcing ordinal consistency along the disease manifold by penalizing score inversions.

\subsection{Directed Diffusion Graph and Regularization}
To capture the local geometry of the disease manifold, we construct a directed weighted graph $\mathcal{G}=(\mathcal{V},\mathcal{E})$ over the samples in a mini-batch. We first define local neighborhoods by feature similarity. For each sample $i$, we compute its $k$-nearest neighbor set $\mathcal{N}_k(i)$, and adopt mutual kNN to keep only pairs that are neighbors of each other, which reduces noisy edges and improves robustness.

We then assign similarity weights using a locally scaled kernel:
\begin{equation}
\widetilde{W}_{ij}=
\exp\left(
-\frac{\|z_i-z_j\|^2}{\sigma_i\sigma_j}
\right),
\end{equation}
where $\sigma_i$ denotes the local scale of sample $i$, which adapts the effective bandwidth to mitigate non-uniform feature density. To explicitly encode the irreversibility prior, we apply a directional mask $\mathbb{I}(y_i\le y_j)$ that keeps edges within the same level or pointing to a more severe level, yielding the directed adjacency matrix
\begin{equation}
W_{ij}=\widetilde{W}_{ij}\cdot \mathbb{I}(y_i\le y_j).
\end{equation}
This construction removes reverse edges when $y_i>y_j$, which prevents the regularizer from enforcing constraints that contradict the progression direction. Since the directional mask can yield rows with $\sum_j W_{ij} = 0$, we add an identity self-loop $W \leftarrow W + I$ to guarantee nonzero row sums and thus a valid random-walk normalization.

Finally, we row-normalize $W$ to obtain a directed transition matrix:
\begin{equation}
P_{ij}=\frac{W_{ij}}{\sum_k W_{ik}},
\end{equation}
where $P_{ij}$ is the probability of diffusing from sample $i$ to sample $j$ along the progression direction.

A one-hop neighborhood may not capture longer progression chains, such as $i\rightarrow m\rightarrow j$. We therefore model multi-scale diffusion via $t$-step random walks~\cite{coifman2006diffusionmaps}:
\begin{equation}
P^{(t)}=P^t,\quad t\in\{1,2,3\}.
\end{equation}

Because every edge respects the non-decreasing severity constraint, $P^{(t)}_{ij}>0$ implies that $j$ is reachable from $i$ through a valid path and thus implicitly indicates $y_i\le y_j$. Multi-scale diffusion propagates local ordinal structure to a broader range, which strengthens global consistency of the regularization.

The key principle of D-ODR is that if sample $i$ can reach sample $j$ through diffusion, the graph construction enforces $y_i \le y_j$, and the predictions should satisfy $s_i \le s_j$. A violation $s_i > s_j$ indicates a reverse flow along the disease manifold and should be penalized. We define D-ODR as

\begin{equation}
\mathcal{L}_{\text{ODR}}
=
\sum_{t\in\mathcal{T}}
\frac{1}{t}
\sum_{i,j}
P^{(t)}_{ij}\,\mathrm{ReLU}\{s_i - s_j\},
\end{equation}

where $\mathrm{ReLU}\{x\}=\max\{x,0\}$ penalizes only ordering violations. The step-decayed weighting emphasizes local consistency while preventing overly strong constraints from long-range diffusion.

\subsection{Supervision, Objective, and Inference}
To learn continuous severity scores, we use ordinal supervision in a regression form with mean squared error (MSE):
\begin{equation}
\mathcal{L}_{\text{MSE}}
=
\frac{1}{N}\sum_{i=1}^{N}(s_i-y_i)^2.
\end{equation}
The overall training objective is a weighted sum of MSE loss and D-ODR:
\begin{equation}
\mathcal{L}
=
\mathcal{L}_{\text{MSE}}
+
\lambda \mathcal{L}_{\text{ODR}},
\end{equation}
where $\lambda$ controls the regularization strength.

At inference time, we compute $s=f_\theta(x)$ only. The D-ODR module is used solely for regularization during training and is not involved in testing, so it introduces no additional inference overhead and keeps $O(1)$ inference complexity.

\section{Datasets and Implementation Details}

\paragraph{Datasets.}
We evaluate our method on four public DR datasets: APTOS~\cite{aptos2019}, Messidor~\cite{abramoff2013messidor,decenciere2014messidor2}, DeepDR~\cite{liu2022deepdr}, and DDR~\cite{li2019ddr}. All datasets are preprocessed and split according to the official GDRBench~\cite{che2023gdrg} protocols.

\paragraph{Implementation Details.}
All experiments are conducted on four NVIDIA RTX A6000 GPUs with a batch size of 128. The initial learning rate is set to $1\times10^{-4}$ and linearly decayed to zero. All images are resized and normalized using standard ImageNet statistics, and random horizontal flipping is applied as the only stochastic data augmentation during training. To ensure statistical reliability, each experiment is repeated three times with different random seeds, and we report the mean performance alongside the standard deviation.

\paragraph{Evaluation.}
Performance is evaluated using the Quadratic Weighted Kappa (QWK) and macro F1 score. We adopt a unified checkpoint selection protocol for all models. For each training run, we compute a composite score by summing the min-max normalized QWK and F1 on the validation set. The checkpoint corresponding to the highest composite score is selected for report.

\section{Results}
\subsection{Comparison with State-of-the-Art Methods}

\begin{table}[h]
\centering
\caption{Performance comparison on four datasets. $\dagger$ and $\ddagger$ indicate ViT-B/16 and RETFound backbones, respectively.}
\label{tab:comparison}

\setlength{\tabcolsep}{2.0pt}

\resizebox{\textwidth}{!}{
\begin{tabular}{l|cc|cc|cc|cc}
\toprule
\multirow{2}{*}{\textbf{Model}} 
& \multicolumn{2}{c|}{\textbf{APTOS}} 
& \multicolumn{2}{c|}{\textbf{Messidor}} 
& \multicolumn{2}{c|}{\textbf{DeepDR}} 
& \multicolumn{2}{c}{\textbf{DDR}} \\
\cmidrule(lr){2-3} 
\cmidrule(lr){4-5} 
\cmidrule(lr){6-7} 
\cmidrule(lr){8-9}
& \textit{QWK} & \textit{F1}
& \textit{QWK} & \textit{F1}
& \textit{QWK} & \textit{F1}
& \textit{QWK} & \textit{F1} \\
\midrule

ViT-B/16\cite{vaswani2017attention} & 87.6{$\pm$1.1} & 65.8{$\pm$1.7} & 57.4{$\pm$2.3} & 51.2{$\pm$3.8} & 74.4{$\pm$0.5} & 56.6{$\pm$0.4} & 75.5{$\pm$1.0} & 54.2{$\pm$0.3} \\
RETFound\cite{zhou2023retfound} & 92.1{$\pm$0.2} & 72.3{$\pm$0.5} & 81.2{$\pm$0.8} & 70.5{$\pm$0.4} & 81.6{$\pm$1.2} & 62.8{$\pm$0.5} & 85.5{$\pm$0.5} & 60.2{$\pm$1.7} \\
\midrule

CORN\textsuperscript{$\dagger$}\cite{shi2023corn} & 88.9{$\pm$0.2} & 68.8{$\pm$0.7} & 60.9{$\pm$2.3} & 49.2{$\pm$5.0} & 74.6{$\pm$2.4} & 56.8{$\pm$0.6} & 73.5{$\pm$1.8} & 53.8{$\pm$2.4} \\
POEs\textsuperscript{$\dagger$}\cite{li2021poes} & 89.6{$\pm$0.7} & 67.4{$\pm$0.5} & 62.7{$\pm$1.1} & 48.1{$\pm$4.3} & 78.8{$\pm$1.4} & 56.4{$\pm$1.3} & 74.8{$\pm$0.8} & 52.4{$\pm$1.0} \\
GOL\textsuperscript{$\dagger$}\cite{lee2022gol} & 88.3{$\pm$1.1} & 64.6{$\pm$1.9} & 52.6{$\pm$2.9} & 45.6{$\pm$1.8} & 74.9{$\pm$1.2} & 51.3{$\pm$0.5} & 75.3{$\pm$1.5} & 50.7{$\pm$0.3} \\
Ord2Seq\textsuperscript{$\dagger$}\cite{wang2023ord2seq} & 89.0{$\pm$0.5} & 67.6{$\pm$0.9} & 60.3{$\pm$2.1} & 50.0{$\pm$6.0} & 75.7{$\pm$1.2} & 56.9{$\pm$1.0} & 75.0{$\pm$1.1} & 54.1{$\pm$0.7} \\
\midrule

CLIP-DR\textsuperscript{$\dagger$}\cite{yu2024clipdr} & 86.9{$\pm$0.4} & 62.8{$\pm$1.1} & 50.7{$\pm$1.6} & 40.9{$\pm$6.5} & 72.9{$\pm$0.6} & 51.3{$\pm$0.6} & 73.7{$\pm$2.0} & 50.4{$\pm$1.0} \\
AOR-DR\textsuperscript{$\dagger$}\cite{yu2025aordr} & 89.4{$\pm$0.5} & 67.2{$\pm$0.7} & 58.2{$\pm$0.9} & 51.5{$\pm$4.0} & 75.0{$\pm$1.1} & 56.3{$\pm$1.6} & 75.3{$\pm$0.7} & 52.1{$\pm$0.7} \\
AOR-DR\textsuperscript{$\ddagger$}\cite{yu2025aordr} & 85.9{$\pm$0.7} & 63.6{$\pm$1.3} & 71.0{$\pm$2.2} & 57.9{$\pm$1.0} & 72.0{$\pm$1.0} & 52.7{$\pm$0.6} & 77.5{$\pm$0.9} & 53.3{$\pm$1.0} \\
\midrule

\rowcolor{lightgray}
\textbf{D-ODR\textsuperscript{$\dagger$}} & 89.8{$\pm$0.8} & 67.0{$\pm$2.0} & 68.2{$\pm$1.4} & 53.0{$\pm$2.4} & 80.1{$\pm$0.7} & 57.4{$\pm$1.5} & 76.8{$\pm$0.3} & 55.7{$\pm$0.5} \\
\rowcolor{lightgray}
\textbf{D-ODR\textsuperscript{$\ddagger$}} & \textcolor{red}{93.2}{$\pm$0.3} & \textcolor{red}{73.3}{$\pm$0.8} & \textcolor{red}{83.9}{$\pm$0.5} & \textcolor{red}{70.9}{$\pm$1.8} & \textcolor{red}{85.7}{$\pm$0.1} & \textcolor{red}{64.7}{$\pm$0.1} & \textcolor{red}{88.2}{$\pm$0.1} & \textcolor{red}{64.6}{$\pm$1.5} \\

\bottomrule
\end{tabular}
}
\end{table}

\begin{table}[h]
\centering
\caption{
Ablation study of different regularization designs with two backbones.
Dir.: direction constraint;
FC: forward constraint;
MSD: multi-scale diffusion.
}
\label{tab:ablation}

\setlength{\tabcolsep}{2.5pt}

\resizebox{\textwidth}{!}{
\begin{tabular}{l|cc|cc|cc|cc}
\toprule
\multirow{2}{*}{\textbf{Model}} 
& \multicolumn{2}{c|}{\textbf{APTOS}} 
& \multicolumn{2}{c|}{\textbf{Messidor}} 
& \multicolumn{2}{c|}{\textbf{DeepDR}} 
& \multicolumn{2}{c}{\textbf{DDR}} \\

\cmidrule(lr){2-3} 
\cmidrule(lr){4-5} 
\cmidrule(lr){6-7} 
\cmidrule(lr){8-9}

& \textit{QWK} & \textit{F1}
& \textit{QWK} & \textit{F1}
& \textit{QWK} & \textit{F1}
& \textit{QWK} & \textit{F1} \\
\midrule

\multicolumn{9}{l}{\textbf{Backbone: ViT-B/16}} \\

Baseline
& 87.7{$\pm$1.1} & 65.7{$\pm$1.7}
& 57.4{$\pm$2.3} & 51.2{$\pm$3.8}
& 74.4{$\pm$0.5} & 56.6{$\pm$0.4}
& 75.5{$\pm$1.0} & 54.2{$\pm$0.3} \\

w/o Dir.
& 88.7{$\pm$0.7} & 62.5{$\pm$1.2}
& 64.7{$\pm$4.4} & 49.8{$\pm$0.4}
& 79.0{$\pm$1.2} & 54.7{$\pm$0.3}
& 76.2{$\pm$1.3} & 49.0{$\pm$0.3} \\

w/o FC
& 87.7{$\pm$2.2} & 58.5{$\pm$13.2}
& 62.4{$\pm$5.2} & 45.9{$\pm$11.9}
& 77.4{$\pm$3.7} & 51.3{$\pm$13.2}
& 75.0{$\pm$1.0} & 49.2{$\pm$4.0} \\

w/o MSD
& 88.8{$\pm$1.7} & 65.9{$\pm$2.3}
& 63.0{$\pm$2.4} & 46.0{$\pm$8.0}
& 80.1{$\pm$1.7} & 56.9{$\pm$1.7}
& 76.3{$\pm$0.7} & 52.5{$\pm$1.2} \\

\rowcolor{lightgray}
\textbf{D-ODR}
& \textcolor{red}{89.8}{$\pm$0.8} & \textcolor{red}{67.0}{$\pm$2.0}
& \textcolor{red}{68.3}{$\pm$1.4} & \textcolor{red}{53.0}{$\pm$2.4}
& \textcolor{red}{80.1}{$\pm$0.7} & \textcolor{red}{57.4}{$\pm$1.4}
& \textcolor{red}{76.8}{$\pm$0.3} & \textcolor{red}{55.7}{$\pm$0.5} \\

\midrule

\multicolumn{9}{l}{\textbf{Backbone: RETFound}} \\

Baseline
& 92.1{$\pm$0.2} & 72.3{$\pm$0.5}
& 81.2{$\pm$0.8} & 70.5{$\pm$0.4}
& 81.6{$\pm$1.2} & 62.8{$\pm$0.5}
& 85.5{$\pm$0.5} & 60.2{$\pm$1.7} \\

w/o Dir.
& 92.3{$\pm$0.5} & 70.9{$\pm$1.8}
& 82.6{$\pm$2.0} & 69.5{$\pm$1.1}
& 84.5{$\pm$1.2} & 64.0{$\pm$0.3}
& 87.9{$\pm$0.2} & 63.5{$\pm$1.2} \\

w/o FC
& 92.5{$\pm$1.1} & 68.1{$\pm$10.1}
& 81.6{$\pm$1.4} & 64.1{$\pm$9.9}
& 84.6{$\pm$2.5} & 55.5{$\pm$16.7}
& 87.4{$\pm$0.9} & 59.5{$\pm$2.8} \\

w/o MSD
& 93.0{$\pm$0.1} & 72.1{$\pm$0.2}
& 82.9{$\pm$0.8} & 67.7{$\pm$1.3}
& 85.0{$\pm$0.3} & 63.8{$\pm$1.1}
& 87.8{$\pm$0.3} & 62.4{$\pm$0.4} \\

\rowcolor{lightgray}
\textbf{D-ODR}
& \textcolor{red}{93.2}{$\pm$0.3} & \textcolor{red}{73.3}{$\pm$0.8}
& \textcolor{red}{83.9}{$\pm$0.5} & \textcolor{red}{70.9}{$\pm$1.8}
& \textcolor{red}{85.7}{$\pm$0.1} & \textcolor{red}{64.7}{$\pm$0.1}
& \textcolor{red}{88.2}{$\pm$0.1} & \textcolor{red}{64.6}{$\pm$1.5} \\

\bottomrule
\end{tabular}
}
\end{table}

Table~\ref{tab:comparison} shows D-ODR achieves state-of-the-art performance across four benchmarks, proving that injecting explicit progression directionality into ordinal learning provides effective inductive bias for DR grading. With the ViT-B/16 backbone, D-ODR performs best nearly uniformly, showing the proposed regularizer particularly benefits generic vision backbones under standard ordinal supervision. ViT-based baselines mainly encode global rank structure, leaving local neighborhoods directionally ambiguous near adjacent grades. D-ODR resolves this by coupling local feature geometry with directed multi-step diffusion, enhancing transitive consistency. Using the RETFound backbone, D-ODR improves the strong baseline, showing it is backbone-agnostic and complementary to advanced ophthalmic pretraining.

However, gains vary across datasets, reflecting differences in label noise, acquisition protocols, and progression cue strength. Highly heterogeneous datasets or those lacking inter-grade separability benefit most, as directional diffusion suppresses contradictory gradients and stabilizes decision boundaries by enforcing monotonicity along plausible transitions. Conversely, if pretrained representations already capture progression-related structudemonstrating backbone-agnosticity and complementarity withconsistent improvements. These results highlight modeling irreversibility and directed reachability as a practical principle improving clinical reliability across diverse DR datasets.

\subsection{Ablation Study}

Table~\ref{tab:ablation} evaluates each algorithmic component's contribution to performance improvements. Removing the directional graph construction (w/o Dir.) reverts the relational manifold to standard laplacian smoothing. This inappropriately enforces bidirectional proximity between samples of varying severities, fails to encode the irreversible progression prior. Consequently, the network learns a clinically ambiguous representation, pulling mild and advanced cases together and washing out ordinal semantics. Distorting the asymmetric forward constraint (w/o FC) by improperly reversing the expected progression penalty induces extreme training instability, yielding drastically inflated F1 score standard deviations across multiple datasets. 

Restricting regularization to one-hop neighborhoods (w/o MSD) limits representational capacity. Single-step transitions only enforce local smoothness, failing to capture the extended disease evolution sequence. Multi-scale diffusion aligns predictions with clinically plausible progression, establishing transitive ordinal relationships. Results across multiple public datasets show consistent gains on complementary metrics, indicating progression-constrained diffusion regularization effectively captures the progressive nature of retinal degradation.

The observed ablation trends remain consistent across both the general-purpose vision transformer and the ophthalmic foundational model. Consistent degradation upon removing any single module confirms conventional representation learning paradigms struggle to preserve strict temporal irreversibility. The proposed components synergistically correct this deficiency, transforming a static ordinal ranking objective into a continuous, biologically plausible geometric flow.

\subsection{Sensitivity Analysis}
\begin{figure}
    \centering
    \includegraphics[width=0.9\linewidth]{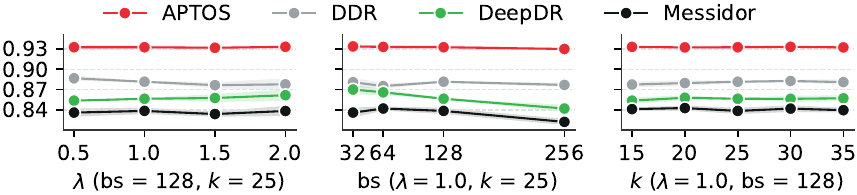}
    \caption{Sensitivity analysis of hyperparameters $\lambda$, $bs$, and $k$ on four datasets. The translucent shaded regions indicate $\pm$ one standard deviation.}
    \label{fig:sa}
\end{figure}

Fig.~\ref{fig:sa} analyzes the sensitivity of the regularization weight $\lambda$, batch size ($bs$), and neighborhood size $k$. The model demonstrates strong robustness to $\lambda$ across the range $[0.5, 2.0]$ and remains largely insensitive to variations in $k$ between 15 and 35. However, while APTOS and DDR are stable across all batch sizes, performance on Messidor and DeepDR degrades noticeably at $bs = 256$, suggesting that excessively large batches may dilute the
local manifold structure essential for diffusion. Based on these observations, we adopt $\lambda = 1.0$, $bs = 128$, and $k = 25$ as the default configuration.

\subsection{Qualitative Analysis}
\begin{figure}
    \centering
    \includegraphics[width=1\linewidth]{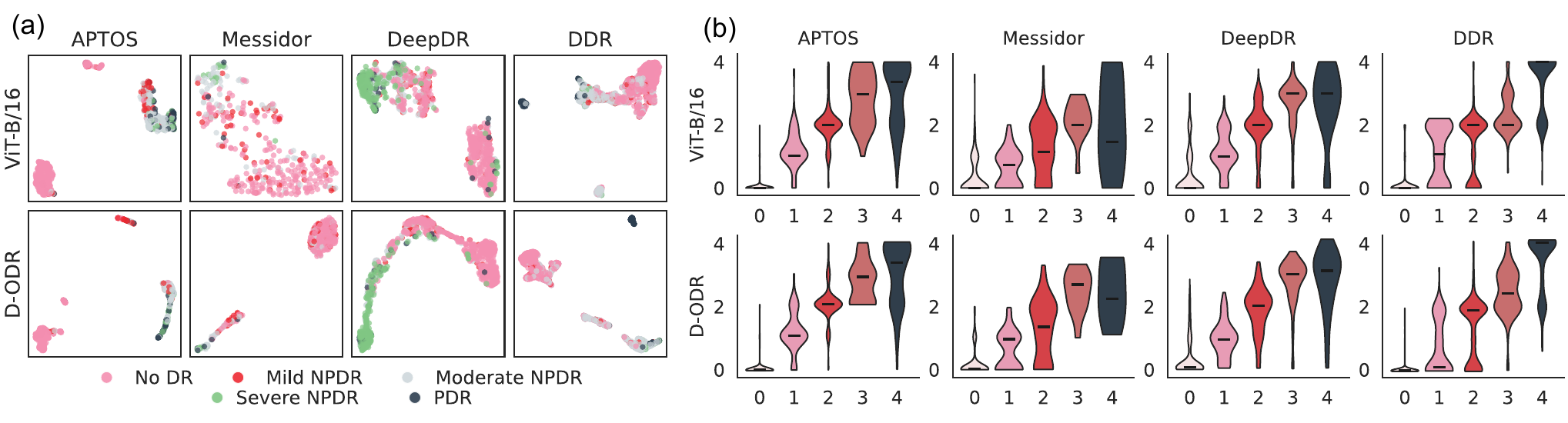}
    \caption{Qualitative evaluation of feature representations and score predictions. (a) 2D UMAP projections; (b) Predicted severity score distributions.}
    \label{fig:qa}
\end{figure}

To intuitively illustrate the representational improvements introduced by D-ODR, Fig.~\ref{fig:qa}(a) visualizes the latent feature space via UMAP projections. The baseline ViT-B/16 yields a fragmented, heavily entangled topology, treating severity grades as isolated clusters and failing to capture the continuous etiology of diabetic retinopathy. In contrast, D-ODR sculpts the latent space into a contiguous, unidirectional manifold, where stages unfold along a clear progressive trajectory from healthy to advanced states. This provides compelling visual evidence that the directed diffusion mechanism embeds the biological irreversibility of DR progression directly into the representation space geometry.

Furthermore, Fig.~\ref{fig:qa}(b) shows the distributions of predicted continuous severity scores. Under the baseline model, intermediate stages exhibit notable intra-class variance and inter-class overlap, highlighting ambiguity near clinical decision boundaries. Conversely, D-ODR refines these distributions, yielding severity intervals that better follow ordinal monotonicity, with median predictions progressing more consistently across stages. While some overlap remains between adjacent grades, improved inter-class alignment and slightly reduced intra-class variance suggest that eliminating contradictory backward gradients during training enables more consistent, biologically faithful severity grading.

\section{Conclusion}

This work proposes D-ODR, a progression-aware ordinal learning framework that incorporates the irreversibility of diabetic retinopathy progression into training. D-ODR constructs a directed diffusion structure and enforces ordinal consistency only along forward trajectories, which aligns predictions with clinically plausible progression and avoids constraints induced by reverse or label-inconsistent relations. Experiments on multiple public datasets show consistent gains over existing methods across complementary metrics, indicating that progression-constrained diffusion regularization provides an effective and backbone-agnostic way to improve continuous severity estimation.

%
% ---- Bibliography ----
%
% BibTeX users should specify bibliography style 'splncs04'.
% References will then be sorted and formatted in the correct style.
%
\bibliographystyle{splncs04}
\bibliography{mybibliography}
%
% \begin{thebibliography}{8}
% \bibitem{ref_article1}
% Author, F.: Article title. Journal \textbf{2}(5), 99--110 (2016)

% \bibitem{ref_lncs1}
% Author, F., Author, S.: Title of a proceedings paper. In: Editor,
% F., Editor, S. (eds.) CONFERENCE 2016, LNCS, vol. 9999, pp. 1--13.
% Springer, Heidelberg (2016). \doi{10.10007/1234567890}

% \bibitem{ref_book1}
% Author, F., Author, S., Author, T.: Book title. 2nd edn. Publisher,
% Location (1999)

% \bibitem{ref_proc1}
% Author, A.-B.: Contribution title. In: 9th International Proceedings
% on Proceedings, pp. 1--2. Publisher, Location (2010)

% \bibitem{ref_url1}
% LNCS Homepage, \url{http://www.springer.com/lncs}, last accessed 2023/10/25
% \end{thebibliography}
\end{document}